\documentclass {article} 
\usepackage{arxiv}
\usepackage{booktabs} 
\usepackage{algorithm}%
\usepackage{algpseudocode}%
\usepackage{hyperref}
\usepackage{amsmath}
\usepackage{xcolor}
\usepackage{xspace}
\usepackage{soul}
\newcommand{\ignore}[1]{}
\usepackage[textsize=tiny]{todonotes}
\usepackage{graphicx}
\usepackage{boldline}
\usepackage[export]{adjustbox}
\usepackage[caption=false]{subfig}
\usepackage[font=small]{caption}
\usepackage{adjustbox}
\usepackage{enumerate}
\usepackage{amsfonts}
\usepackage{colortbl}
\usepackage{float}
\usepackage{rotating}
\usepackage{setspace}
\usepackage{gensymb}	
\usepackage[official]{eurosym} 
\usepackage[flushleft]{threeparttable}
\usepackage{enumitem}
\usepackage{array}
\title{Design optimisation of a multi-mode wave energy converter}

\author{
Nataliia Y. Sergiienko\\
School of Mechanical Engineering\\
	The University of Adelaide\\
	 Australia\\
	 \texttt{nataliia.sergiienko@adelaide.edu.au} \\
\And
  Mehdi Neshat \\
  Optimization and Logistics Group\\
  School of Computer Science\\
  The University of Adelaide\\
   Australia \\
  \texttt{mehdi.neshat@adelaide.edu.au} \\
   \And
   Leandro S.P. da Silva\\
   School of Mechanical Engineering\\
	The University of Adelaide\\
	 Australia\\
   \texttt{leandro.dasilva@adelaide.edu.au}
   \And
 Bradley Alexander \\
  Optimization and Logistics Group\\
  School of Computer Science\\
  The University of Adelaide\\
   Australia \\
  \texttt{bradley.alexander@adelaide.edu.au} \\
  \And
 Markus Wagner \\
  Optimization and Logistics Group\\
  School of Computer Science\\
  The University of Adelaide\\
   Australia \\
  \texttt{markus.wagner@adelaide.edu.au} \\
  }

\begin{document}

\maketitle
\doublespacing
\begin{abstract}
A wave energy converter (WEC) similar to the CETO system developed by Carnegie Clean Energy is considered for design optimisation. This WEC is able to absorb power from heave, surge and pitch motion modes, making the optimisation problem nontrivial. The WEC dynamics is simulated using the spectral-domain model taking into account hydrodynamic forces, viscous drag, and power take-off forces. The design parameters for optimisation include the buoy radius, buoy height, tether inclination angles, and control variables (damping and stiffness). The WEC design is optimised for the wave climate at Albany test site in Western Australia considering unidirectional irregular waves.
Two objective functions are considered: (i) maximisation of the annual average power output, and (ii) minimisation of the levelised cost of energy (LCoE) for a given sea site. The LCoE calculation is approximated as a ratio of the produced energy to the significant mass of the system that includes the mass of the buoy and anchor system. Six different heuristic optimisation methods are applied in order to evaluate and compare the performance of the best known evolutionary algorithms, a swarm intelligence technique and a numerical optimisation approach.
The results demonstrate that if we are interested in maximising energy production without taking into account the cost of manufacturing such a system, the buoy should be built as large as possible (20 m radius and 30 m height). However, if we want the system that produces cheap energy, then the radius of the buoy should be approximately 11-14~m while the height should be as low as possible. These results coincide with the overall design that Carnegie Clean Energy has selected for its CETO 6 multi-moored unit. However, it should be noted that this study is not informed by them, so this can be seen as an independent validation of the design choices.

\end{abstract}


\sloppy

\section{Introduction}

\begin{figure}
\centering
\subfloat[]{
\includegraphics[clip,width=0.3\columnwidth]{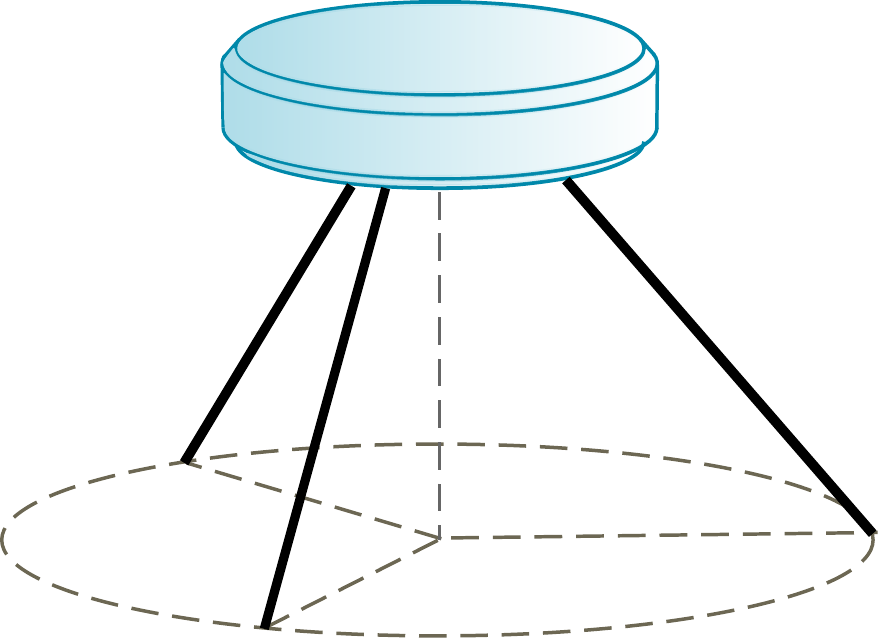}}
\subfloat[]{
\includegraphics[clip,width=0.3\columnwidth]{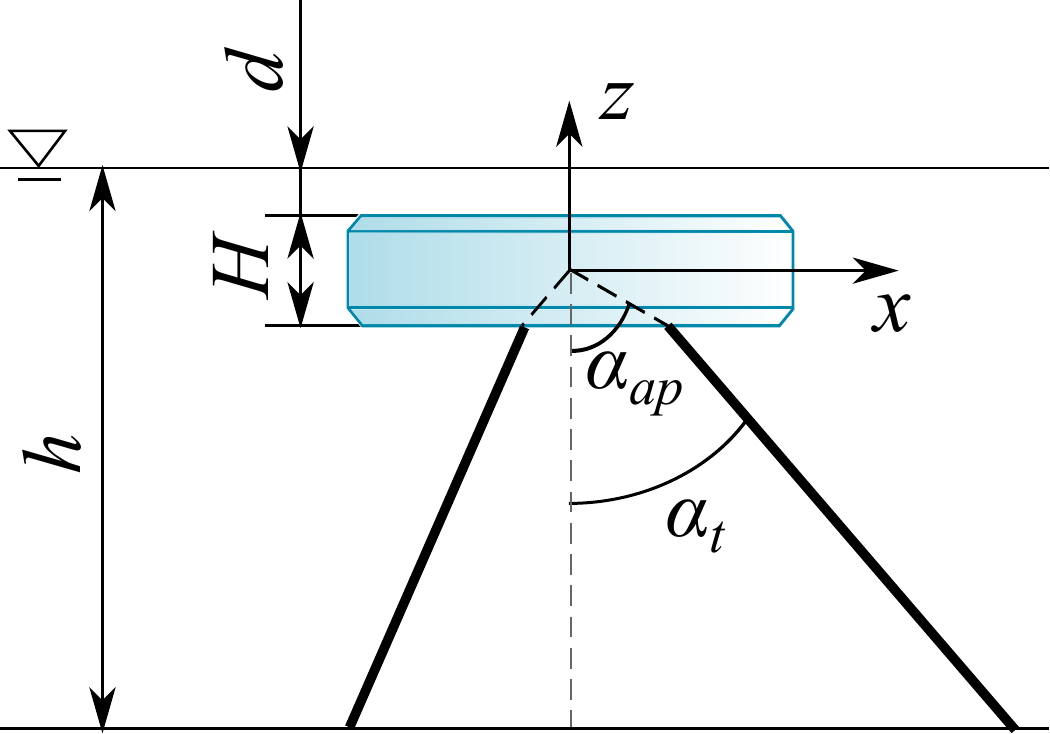}}
\subfloat[]{
\includegraphics[clip,width=0.3\columnwidth]{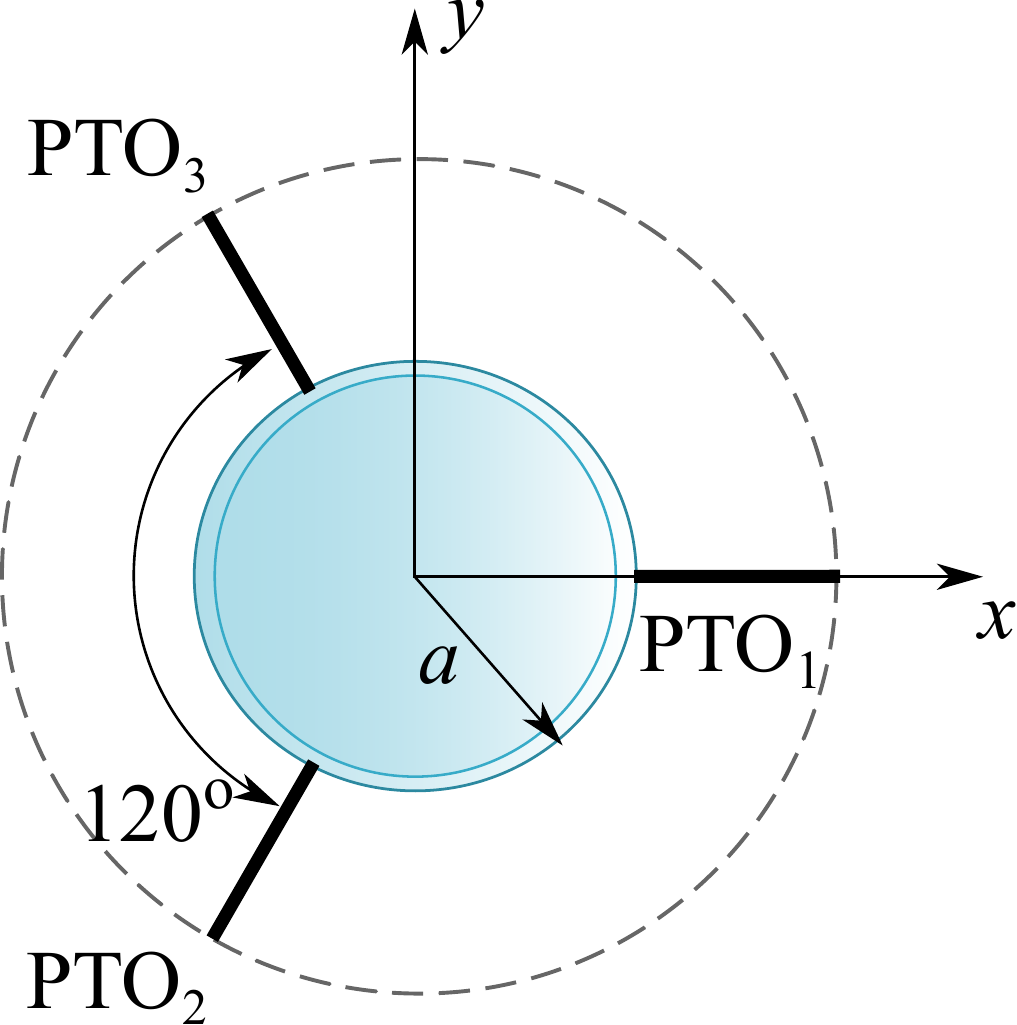}}
\caption{ Geometry and parameters of the three-tether wave energy converter: (a)~3D view, (b)~front view, (c)~top view.}%
\label{fig:buoy}
\end{figure}
The geometry of a wave energy converter (i.e., its shape and size) determines how efficiently it radiates waves and absorbs incident wave power. A number of studies have been conducted to develop recommendations for wave energy developers regarding the hydrodynamic design of WECs \cite{Falnes2012, Hals2013}. In the latter, the optimal dimensions of the wave absorbing body are chosen from the point of view of maximising power production, and depending on the mode of oscillation (heave, surge, or pitch) \cite{Hals2017} and applied control strategy \cite{Garcia2015b}. Similarly, the annual power output has been used as an objective function for shape optimisation of various WECs in \cite{Wen2018, Rezvan2019, Soheil2019, Wang2019}.

However, the economic attractiveness of any wave energy project depends not only on its power production, but also on the costs associated with a project lifetime \cite{Pecher2017}. Therefore, the economic component should be integrated into the WEC design to make wave energy cost-effective. Despite the fact that the levelised cost of energy is the most reliable metric to assess energy investments, its calculations for wave energy devices are full of assumptions and uncertainties. Therefore, other cost-related measures have been widely used for techno-economic development of WECs. Thus, power per displaced volume of the buoy was utilised in \cite{McCabe2013} to optimise the shape of a surging WEC for a site in the North-East Atlantic Ocean. As the optimised shapes were complex and not adequate for construction, the research has been extended in \cite{Garcia-Teruel2019} by including the cost of materials and manufacturing processes into the optimisation procedure. In addition to the costs associated with a buoy construction, \cite{Clark2018, Clark2019} suggest taking into account the PTO-reliability at the early stages of the WEC hull optimisation.

The majority of geometry optimisation studies consider WECs that absorb power from one hydrodynamic mode (e.g. heave \cite{Soheil2019}, surge \cite{Garcia-Teruel2018}, or pitch \cite{Rezvan2019}). This work focuses on the CETO 6 technology being under development by Carnegie Clean Energy Limited, Australia. The CETO 6 design was announced in 2017 and consists of a submerged cylindrical buoy (25~m diameter) attached to three mooring lines capturing power from heave, surge and pitch motions. The first attempts to understand how the dimensions of a submerged cylindrical buoy (radius and height) affect its power efficiency depending on the motion mode has been made in \cite{Sergiienko2016}. However, only a number of prescribed geometries were investigated without a properly designed optimisation procedure. As a result, this paper addresses three main objectives:

\begin{enumerate}[label=(\roman*), topsep = 0pt]
    \item to optimise the geometry of a WEC that absorbs power from surge, heave and pitch simultaneously;
    \item to investigate how this geometry depends on the chosen objective function: power production, or LCoE;
    \item to understand the design solution for CETO 6.
\end{enumerate}

\section*{MODELLING}

\subsection*{Wave energy converter}

The wave energy converter considered in this work is shown in Fig.~\ref{fig:buoy}. A fully submerged cylindrical buoy is tied to the seabed through three tethers. It is assumed that each tether is connected to an individual power take-off (PTO) machinery that is capable to behave as a spring-damper system. The parameters that define the size of the cylinder are radius ($a$), and height ($H$). The buoy is designed to operate at $d = 2$~m below the water surface (from the buoy top to the still water level), in the water column of $h = 50$~m. The attachment of tethers to the buoy hull is defined by two distinct angles: the tether inclination angle ($\alpha_t$), and the tether attachment angle, or attachment point, ($\alpha_{ap}$). The mass of the buoy is equal to  half the displaced mass of water $m_b = 0.5\rho_w V$ ($\rho_w = 1025$~kg/m$^3$, and $V = \pi a^2 H$).

\begin{figure}
    \centering
    \includegraphics[width=0.7\linewidth]{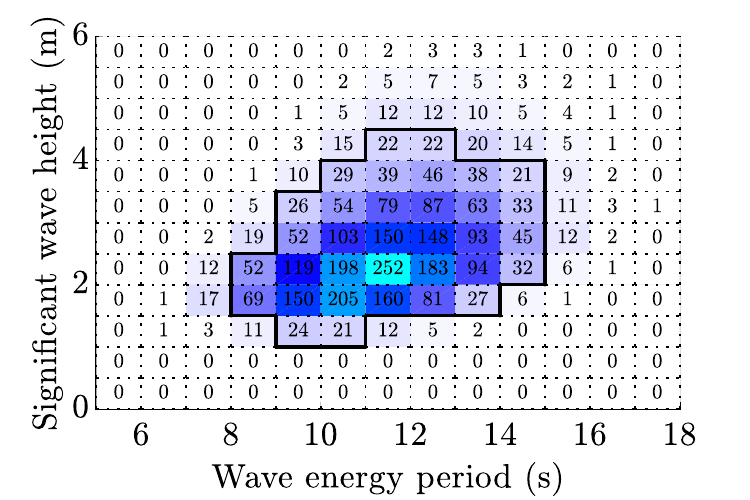}
    \caption{The wave climate at the Albany deployment site located in Western Australia (117.7547\degree E, 35.1081\degree S, 33.9~kW/m mean annual wave power resource) \cite{AWAVEA}.}
    \label{fig:Albany}
\end{figure}

This WEC will be installed at the so-called Albany test site in Western Australia, which has a wave climate specified in Fig.~\ref{fig:Albany}. In order to reduce the number of sea states considered to assess the WEC performance, a sub-set of 34 sea states with a total probability of 99\% are chosen to represent the deployment climate (outlined by a black line).

\subsection*{Time-domain model}

A detailed description of the non-linear time domain model of the three-tether WEC has been documented in \cite{Scruggs2013, Sergiienko2018, Sergiienko2019} and only key points are outlined in the following.

The WEC has six degrees-of-freedom, and its motion is described by the position vector $\mathbf{x}\in \mathbb{R}^{6\times1}$ ($x_1$ surge, $x_2$ sway, $x_3$ heave, $x_4$ roll, $x_5$ pitch, $x_6$ yaw) in the reference (Cartesian) coordinate frame $Oxyz$. The buoy is connected to three tethers that are represented by the vector of tether length variables $\mathbf{q} = [l_1\quad l_2 \quad l_3]^{\operatorname{T}}$. The kinematic relationship between the buoy velocity $\dot{\mathbf{x}}$ and the rate of change of the tether length has a form of $ \dot{\mathbf{q}}(t) = \mathbf{J}^{-1}(\mathbf{x})\dot{\mathbf{x}}(t)$,
where $\mathbf{J}^{-1}(\mathbf{x})\in \mathbb{R}^{3\times6}$ is the inverse kinematic Jacobian which is a function of the buoy current pose \cite{Sergiienko2016}.

The motion of the three-tether WEC in the time domain can be described by the following equation:
\begin{equation}
\mathbf{M}\ddot{\mathbf{x}}(t) = \mathbf{F}_{exc}(t) + \mathbf{F}_{rad}(t) + \mathbf{F}_{visc}(t) + \mathbf{F}_{buoy}(t) + \mathbf{F}_{tens}(t), \label{eq:nonlinear}
\end{equation}
where $\mathbf{M}$ is a mass matrix, $\mathbf{F}_{exc}$ is the wave excitation force, $\mathbf{F}_{rad}$ is the wave radiation force, $\mathbf{F}_{visc}$ is the viscous damping force, $\mathbf{F}_{buoy}$ is the buoyancy force, $\mathbf{F}_{tens}$ is the generalised tether force which includes the initial tension in the tethers that counteracts the buoyancy force, and the control forces exerted on the buoy from the PTO machinery.

As optimisation procedures typically require a large number of evaluations to be performed in order to find the best WEC configuration for a given objective function, a low computational cost is desired: the lower the computational cost of the model, the faster the optimisation algorithm converges in terms of wallclock time. Therefore, in order to develop the computationally efficient model suitable for optimisation purposes, the nonlinear Eq.~\eqref{eq:nonlinear} is replaced by its linear equivalent spectral-domain model using the statistical linearisation technique.

\subsection*{Spectral-domain model}

The spectral-domain model of the system is a set of linear equations of motion written in the frequency domain that approximates the system dynamics by replacing all nonlinear terms with their linear counterparts \cite{Folley2016}. For the proposed three-tether device, the main source of nonlinearity comes from the viscous drag force, while the geometric nonlinearity from the tethers can be linearised around the equilibrium position. Thus, assuming a harmonic response of the WEC $\mathbf{x}(t) = \mathfrak{Re}\{\hat{\mathbf{x}} \; e^{i\omega t}\}$,
the equivalent linear system for the nonlinear Eq.~\eqref{eq:nonlinear} in the frequency domain has a form:
\begin{equation}
    \left[-\omega^2\left(\mathbf{M} + \mathbf{A}\right) + i\omega \left(\mathbf{B} + \mathbf{B}_{pto} + \mathbf{B}_{eq}\right) + \mathbf{K}_{pto}\right]\hat{\mathbf{x}}(\omega) =
    \hat{\mathbf{F}}_{exc}(\omega) \label{eq:sl}
\end{equation}
where $\mathbf{A}(\omega)$ and $\mathbf{B}(\omega)$ are the frequency dependent added mass and radiation damping matrices, $\mathbf{K}_{pto}$ and $\mathbf{B}_{pto}$ are obtained from the linearisation of the generalised tether force $\hat{\mathbf{F}}_{tens}(\omega) = -\mathbf{B}_{pto}\hat{\dot{\mathbf{x}}}(\omega) - \mathbf{K}_{pto}\hat{\mathbf{x}}(\omega)$ (see \cite{Scruggs2013} for more details), and $\mathbf{B}_{eq}$ is the equivalent damping matrix that corresponds to the quadratic damping nonlinearity. The value of $\mathbf{B}_{eq}$ is unknown and can be determined iteratively using the statistical linerisation \cite{Leandro2020}:
\begin{equation}
    \mathbf{B}_{eq} = -\left\langle \frac{\partial \mathbf{F}_{visc}}{\partial \dot{\mathbf{x}}} \right\rangle,
\end{equation}
where $\left\langle \cdot \right\rangle$ denotes mathematical expectation, and the viscous force is defined as:
\begin{equation}
    \mathbf{F}_{visc} = -\frac{1}{2}\rho_w \mathbf{C}_d \mathbf{A}_d (||\dot{\mathbf{x}}||\odot \dot{\mathbf{x}}),
\end{equation}
$\mathbf{C}_d$ and $\mathbf{A}_d$ are the matrices of the drag coefficients and the cross-section areas of the buoy perpendicular to the direction of motion
respectively.

To calculate $\mathbf{B}_{eq}$ and estimate the approximate response of the WEC in irregular waves, the following iterative procedure is employed:
\begin{enumerate}[label=Step \arabic*., wide=0pt, leftmargin=*]
    \item Define the incident wave spectrum $S_{\eta}(\omega)$
    \item Calculate the power spectral density (PSD) matrix of the excitation force:
    \begin{equation}
        \mathbf{S}_\mathbf{F}(\omega) = S_{\eta}(\omega) \cdot \mathrm{diag} (|\hat{f}_{exc,1}|^2, \ldots , |\hat{f}_{exc,6}|^2)
    \end{equation}
    where $\hat{f}_{exc,i}(\omega)$ is the frequency-dependent excitation force coefficient for the $i$-th degree of freedom.
    \item Obtain the frequency response matrix of the WEC:
    \begin{equation}
        \mathbf{H}(\omega) = \left[-\omega^2\left(\mathbf{M} + \mathbf{A}\right) + i\omega \left(\mathbf{B} + \mathbf{B}_{pto} + \mathbf{B}_{eq}\right) + \mathbf{K}_{pto}\right]^{-1}
    \end{equation}
    assuming $\mathbf{B}_{eq}=\mathbf{0}_{6\times 6}$ in the first iteration.
    \item Determine the PSD matrix of the buoy response:
     \begin{equation}
        \mathbf{S}_{\mathbf{x}}(\omega) = \mathbf{H}(\omega)\mathbf{S}_\mathbf{F}(\omega)\mathbf{H}^{\rm T*}(\omega).
    \end{equation}
    \item Calculate the variance of the WEC velocity in each degree of freedom $i = 1\ldots 6$:
    \begin{equation}
        \sigma_{\dot{x}_i}^2 = \int_{0}^{\infty}\omega^2 S_{x_i}(\omega) d\omega
    \end{equation}
    \item Estimate the equivalent damping matrix $\mathbf{B}_{eq} = \operatorname{diag}(B_{eq,1}, \ldots, B_{eq,6})$, where according to \cite{Leandro2020}:
    \begin{align}
        B_{eq_i} = -\left\langle \frac{\partial F_{visc_i}}{\partial \dot{x}_i} \right\rangle = \frac{1}{2}\rho_w C_{d_i} A_{d_i} \sqrt{\frac{8}{\pi}}\sigma_{\dot{x}_i}
    \end{align}
    \item Check the convergence criteria: the difference between all elements of $\mathbf{B}_{eq}$ obtained in the current $[n]$ and previous $[n-1]$ iterations should be less then a threshold $\delta = 0.01$:
    \begin{equation}
       |\mathbf{B}_{eq}[n] - \mathbf{B}_{eq}[n-1]|<\delta.
    \end{equation}
    If the convergence is not achieved, go to Step 3.
\end{enumerate}

It usually takes up to 10 iterations to find the equivalent damping matrix $\mathbf{B}_{eq}$ for any WEC geometry operating in one sea state.

The average power output from each of the three power take-off units $k = 1\ldots 3$ can be found as \cite{Leandro2020}:
\begin{equation}
    \bar{P}_k = 
    B_{pto} \sigma^2_{\dot{q}_k},
\end{equation}
where $\sigma^2_{\dot{q}_k}$ is the variance of the tether length rate change $\dot{\mathbf{q}}$ that can be calculated as:
\begin{equation}
    \sigma^2_{\dot{q}_k} = \int_{0}^{\infty}\omega^2 S_{q_k}(\omega) d\omega,
\end{equation}
where $\mathbf{S}_{\mathbf{q}}(\omega) = \mathbf{J}^{-1}_0 \mathbf{S}_{\mathbf{x}}(\omega)\mathbf{J}^{-\operatorname{T}}_0$, and $\mathbf{J}^{-1}_0 = \mathbf{J}^{-1} (\mathbf{x}_0)$ is the inverse kinematic Jacobian at the nominal position of the buoy $\mathbf{x}_0 = \mathbf{0}_{6\times 1}$.

Thus, the power generated by the WEC in a sea state characterised by the significant wave height $H_s$ and peak wave period $T_p$ is:
\begin{equation}
    \bar{P}(H_s, T_p) = B_{pto} \sum_{k = 1}^{3}\sigma^2_{\dot{q}_k}(H_s, T_p).
\end{equation}

If the probability of occurrence of each sea state ($H_s, T_p$) is characterised by the matrix $\mathbf{O}(H_s, T_p)$, then the total average annual power output generated by the WEC in the wave climate can be calculated as:
\begin{equation}
    P_{AAP} = \sum_{H_s} \sum_{T_p} O(H_s, T_p)\cdot \bar{P}(H_s, T_p). \label{eq:Paap}
\end{equation}

\begin{figure}[hbt]
	\centering
	\includegraphics[width=0.5\linewidth]{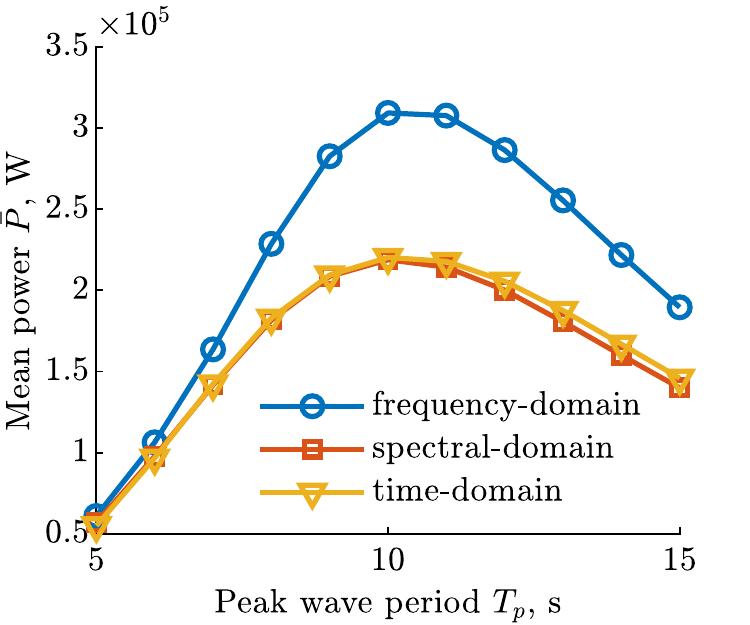}
	\caption{Average power output of the three-tether WEC ($a = 5.5$~m, $H = 5.5$~m, $\alpha_{ap} = \alpha_t = 45 \deg$, $K_{pto} = 200$~kN/m, $B_{pto} = 150$~kN/(m/s)) calculated using frequency-, time- and spectral domain models. All sea states have the significant wave height of $H_s = 3$~m and are described by the Pierson-Moskowitz spectrum.}\label{fig:Power}
\end{figure}

To demonstrate the effectiveness of the spectral-domain model,
the average output power of the three-tether WEC is shown in Fig.~\ref{fig:Power} using three different models (frequency, spectral, and time domain). As expected, the frequency domain model (Eq.~\eqref{eq:sl} assuming $\mathbf{B}_{eq} = \mathbf{0}$) significantly overestimates the absorbed power. The time domain and spectral domain models provide similar estimates of power output, while the latter is approximately several orders of magnitude faster.

\begin{figure}[hbt]
	\centering
	\includegraphics[width=0.5\linewidth]{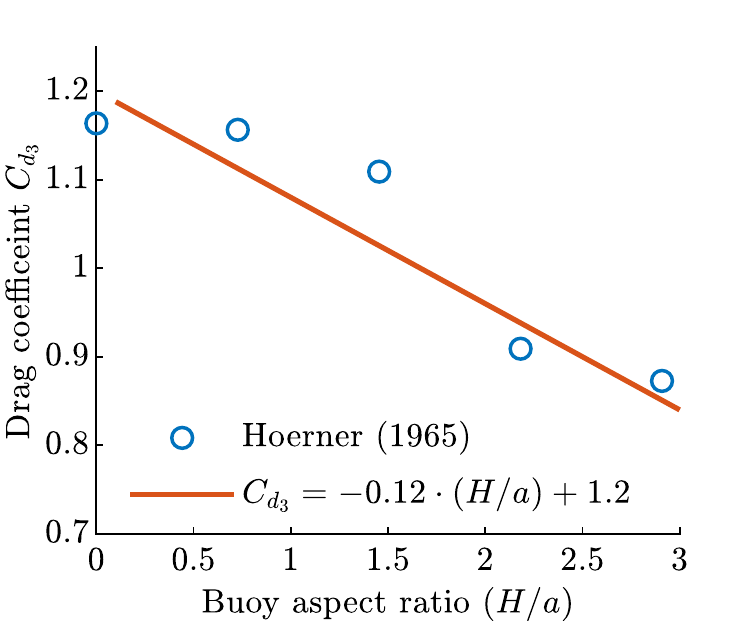}
	\caption{Drag coefficient of the cylindrical body in axial flow as a function of its aspect ratio $H/a$.}\label{fig:drag}
\end{figure}

The values of the viscous drag force and, consequently, the values of the linearised $\mathbf{B}_{eq}$ matrix depend on the drag coefficients $C_{d_i}$ ($i = 1\ldots 6$) taken for the analysis. $C_{d_i}$ depends on the ratio of the cylinder length (or height) to its diameter, especially for the heave direction $i = 3$. Therefore, for optimisation purposes, the drag coefficient in heave $C_{d_3}$ is represented as a function of the buoy aspect ratio ($H/a$) based on data from \cite{drag1965}, see Fig.~\ref{fig:drag}. So $C_{d_3} = -0.12(H/a) + 1.2$, while for other motion modes, drag coefficients are kept fixed regardless of the cylinder aspect ratio: $C_{d_1} = C_{d_2} = 1$ for surge and sway, and $C_{d_4} = C_{d_5} = 0.2$ for roll and pitch.

\subsection*{Approximation of LCoE value}

Due to the lack of commercial wave power plants, the calculations of the actual LCoE value are based on many uncertainties associated with power production and related costs. \cite{Andres2016} assessed various techno-economic related metrics and made a conclusion, that for the CorPower device LCoE can be approximated by the following equation: 

\begin{equation}
    \textrm{LCOE} \left(\frac{\textrm{\euro}}{\textrm{kWh}} \right) = \textrm{RDC} \times \left(\frac{\textrm{Energy (MWh)}}{\textrm{Mass (kg)}} \right)^{-0.5},
\end{equation}
where the mass corresponds to a significant mass of the WEC including the mass of the buoy and the anchoring system, and RDC is a site-dependent coefficient that will be omitted in calculations (RDC = 1) because optimisation is done for the same deployment site.

In this study, the calculation of the significant mass of the WEC is based on the following assumptions:
\begin{itemize}
    \item[-] the buoy mass is $m_b = 0.5\rho_w \pi a^2 H$;
    \item[-] the required mass of the anchoring system (three piles) depends on the tether tension associated with buoyancy and the wave force. Therefore, it can be approximated by:
    \begin{equation}
        m_{as} = \frac{m_{as}^{\textrm{ref}}}{(F_{t}^{peak})^{\textrm{ref}}}F_{t}^{peak},
    \end{equation}
    where a three-tether WEC with a $a = 5.5$~m radius and $H = 5.5$~m height is taken as a reference case from \cite{Sergiienko2018} (the mass of three piles is estimated as $m_{as}^{\textrm{ref}} = 225\times 10^3$~kg, and the statistical peak force is $(F_{t}^{peak})^{\textrm{ref}} = 1.94$~MN). Therefore, in order to estimate $m_{as}$ for any given cylinder dimensions, it is required to calculate the tether peak force ($99\% = 2.57\sigma_{F_{t}}$) using the spectral domain model.
\end{itemize}
As a result, the LCoE model used in this study is:
\begin{equation}
    \textrm{LCOE} = \left(\frac{8760 P_{AAP}}{m_b + m_{as}} \right)^{-0.5}. \label{eq:lcoe}
\end{equation}


\section*{OPTIMISATION ROUTINE}

\subsection*{Objective functions}

The parameters of the WEC that are considered for optimisation are: the buoy radius $a$, the buoy aspect ratio defined as the ratio of the buoy height to its radius ($H/a$), the tether inclination angle $\alpha_{t}$, the tether attachment angle $\alpha_{ap}$, the vector of PTO stiffness coefficients for each of the $N=34$ sea state considered $\mathbf{k}_{pto} = [K_{pto}^{(1)}, \; K_{pto}^{(2)}, \ldots, K_{pto}^{(N)}]^{\operatorname{T}}$, the vector of PTO damping coefficients $\mathbf{b}_{pto} = [B_{pto}^{(1)}, \; B_{pto}^{(2)}, \ldots, B_{pto}^{(N)}]^{\operatorname{T}}$. In total, there are 72 parameters that should be optimised:
\begin{equation}
    \mathbf{z} = [a,\; (H/a), \; \alpha_t, \; \alpha_{ap},\; \mathbf{k}_{pto}\in \mathbb {R}^{N\times 1}, \; \mathbf{b}_{pto} \in \mathbb {R}^{N\times 1}].
\end{equation}

The two objective functions considered in this study are:
\begin{enumerate}[label = (\roman*), topsep = 0pt]
    \item maximise average annual power output calculated using Eq.~\eqref{eq:Paap}:
    \begin{equation}
        f_{O1} = \underset{\mathbf{z}}{\arg\max}\; P_{AAP}(\mathbf{z}), \, \textrm{subject to: } \mathbf{z}\in[\mathbf{z}_{\min}, \mathbf{z}_{\max}]
    \end{equation}
    \item minimise LCoE value specified in Eq.~\eqref{eq:lcoe}:
    \begin{equation}
        f_{O2}  = \underset{\mathbf{z}}{\arg\min}\; \textrm{LCOE}(\mathbf{z}), \, \textrm{subject to: } \mathbf{z}\in[\mathbf{z}_{\min}, \mathbf{z}_{\max}]
\end{equation}
\end{enumerate}

The range of the design and PTO parameters used in optimisation is specified in Table~\ref{tab:constraints}.

\begin{table}[h]
\small
    \caption{Constraints on the design parameters.}
    \label{tab:constraints}
    \centering
    \setlength\extrarowheight{-3pt}
    \begin{tabular}{lccc}
    \toprule
        Parameter & Unit & Min & Max\\
        \midrule
        Buoy radius, $a$ & m & 5 & 20\\
        Buoy aspect ratio, $(H/a)$ & & 0.4 & 1.5\\
        Tether inclination angle, $\alpha_{t}$ &deg & 10 & 80\\
        Tether attachment angle, $\alpha_{ap}$ &deg & 10 & 80\\
        PTO stiffness, $K_{pto}$ & N/m & $10^3$ & $10^8$\\
        PTO damping, $B_{pto}$ & N/(m/s) & $10^3$ & $10^8$\\
        \bottomrule
    \end{tabular}
\end{table}
\subsection*{Optimisation methods}


All methods that we consider are iterative, heuristic approaches, because we do not have a closed mathematical form that might make the optimisation amenable to efficient, purely mathematical optimisation. We employ a range of structurally different methods, not only with the goal of comparing their performance, but also in order to mitigate the inherent search bias that each heuristic has. The latter has the advantage that potentially different locally optimal designs can be compared.

The six methods that we evaluate are:

\begin{enumerate}[topsep=0pt, itemsep=0pt]
    \item Nelder-Mead (NM) \cite{lagarias1998convergence}, which is a deterministic hill-climber for problems where derivatives are unknown;
    \item a simple evolutionary algorithm (1+1EA)~\cite{eiben2007parameter}, which is a stochastic hill-climber, and which (in each iteration) mutates each parameter of a WEC design with a probability of $\displaystyle \frac{1}{\textit{number of design variables}}$ using a normal distribution ($\sigma=0.1\times(U_b-L_b)$), where $U_b$ and $L_b$ denote the respective variable's upper and lower bound;
    \item Particle Swarm Optimisation (PSO)~\cite{eberhart1995new}, which is a popular heuristic, and which we use with $\lambda=25$, $c_1=1.5$, $c_2=2$, $\omega=1$ (which is exponentially decreased with a damping ratio $w_f=0.99$); 
    \item Covariance Matrix Adaptation Evolution Strategy (CMA-ES)~\cite{hansen2006cma}, which is a state-of-the-art self-adaptive optimiser for continuous spaces, and which we use with the default settings and  $\lambda=16$;
    \item Differential Evolution (DE)~\cite{storn1997differential}, which is a  state-of-the-art method from a different class of algorithms, which employs the concept of crossover to recombine information from other designs, and which we use with a configuration of $\lambda=25$ (population size), $F=0.5$ and $P_{cr}=0.8$; 
    \item SaDE \cite{qin2008differential}, which is a self-adaptive differential evolution. Every one of the strategies/operators in the pool initially has the same probability of being applied. During the evolution, the probabilities are updated based on the number of successfully generated offspring. 
\end{enumerate}

Each optimisation run is given a computational budget of 5000 evaluations. 
As the methods are randomised iterative approaches\footnote{NM randomly samples the initial WEC designs}, we repeat each optimisation 10 times.

\section*{RESULTS AND DISCUSSION}

\subsection*{Objective function $f_{O1}$}

\begin{figure}[hbt]
	\centering
	\includegraphics[width=0.5\linewidth]{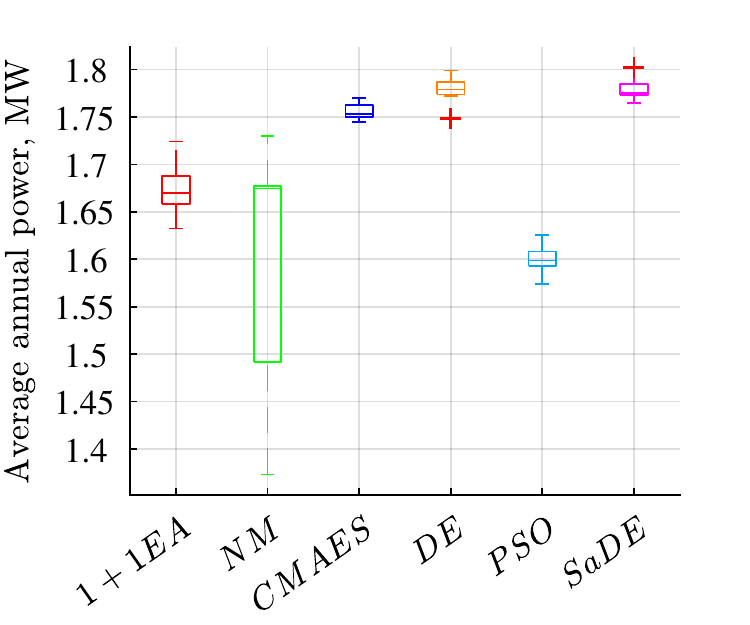}
	\caption{Performance comparison of various optimisation algorithms using objective function $f_{O1}$ (10 independent runs per each method).}\label{fig:max_power}
\end{figure}

This objective function is related to the maximisation of the annual average power output. The power outputs of the best WEC designs for all six optimisation algorithms are shown as box and whisker plots in Fig.~\ref{fig:max_power}. The corresponding optimised design parameters are presented in Table~\ref{tab:max_power}. 

\begin{table}[htb]
    \small
    \setlength\extrarowheight{-3pt}
    \caption{Optimised design parameters when average annual power is maximised.}
    \label{tab:max_power}
    \centering
    \begin{tabular}{ccccccc}
        \toprule
       Parameter & 1+1EA & NM & CMA-ES & DE & PSO & SaDE \\
      \midrule
       $a$ [m]          & 18.8  & 18  & 19.8  & 19.1  & 18    & 19\\
       $H/a$          & 1.5   & 1.49  & 1.5   & 1.5   & 1.5   & 1.5\\
       $\alpha_t$ [deg]  & 31    & 26    & 28    & 28    & 33    & 26\\
       $\alpha_{ap}$ [deg] & 56   & 55    & 53    & 54    & 53    & 53\\
       \midrule
       $P_{AAP}$  [MW]  & 1.72  & 1.73  & 1.77  & 1.77 & 1.63 & 1.8 \\
      \bottomrule
    \end{tabular}
\end{table}

It can be observed that almost all algorithms converged to the configurations with a maximum (or close to) possible radius and height ($a = 18-20$~m, $H/a = 1.5$, so $H = 27-30$~m), with the highest power output of 1.8 MW produced by SaDE.
The radius of the cylinder mainly affects the power absorption properties of the WEC in heave, while its height affects power from surge. As a result, in order to maximise the power production from each mode of oscillation, a cylindrical WEC should be built as large (and tall) as possible. These results are in agreement with \cite{BabaritReport2011}, where the power increases with the buoy volume. However, it should be noted that the growth rate of power decreases with the increase in the buoy size, and the cost of manufacturing such large WECs might be not justified by the power generation.

The optimal value of the tether angle $\alpha_t$ depends on the hydrodynamic mode (heave or surge) that is dominant in power production. Thus, for the heaving configuration as $H/a \to 0$, $\alpha_t \to 0$~deg (should converge to a single-tether WEC), while for the surging case as $H/a \to \infty$, $\alpha_t \to 90$~deg. According to \cite{Sergiienko2016}, the expected value of $\alpha_t$ for $H/a = 1.5$ should be approximately 52 deg, while all optimisation algorithms provided values $\alpha_t\approx 30$~deg. There might be several reasons for these results. First of all, modelling in \cite{Sergiienko2016} has been done using a linear frequency domain model, while in this study the model includes the viscous drag force. Secondly, $\alpha_t$, $\alpha_{ap}$, $K_{pto}$ and $B_{pto}$ have a coupled effect on the power generation, and it is possible that there are several combinations of their optimal values that lead to a similar power output.

The optimised values of the tether attachment angle $\alpha_{ap}$ are within a range of 53-56 deg. This angle affects the controllability and power output from pitch motion, so because $\alpha_{ap}^{\textrm{opt}}>\alpha_t^{\textrm{opt}}$ the tethers provide better control over pitch. However, we are not aware of any studies on what values should be expected for this cylinder configuration.

\begin{figure}[htb]
    \centering
    \includegraphics [width=0.75\textwidth]{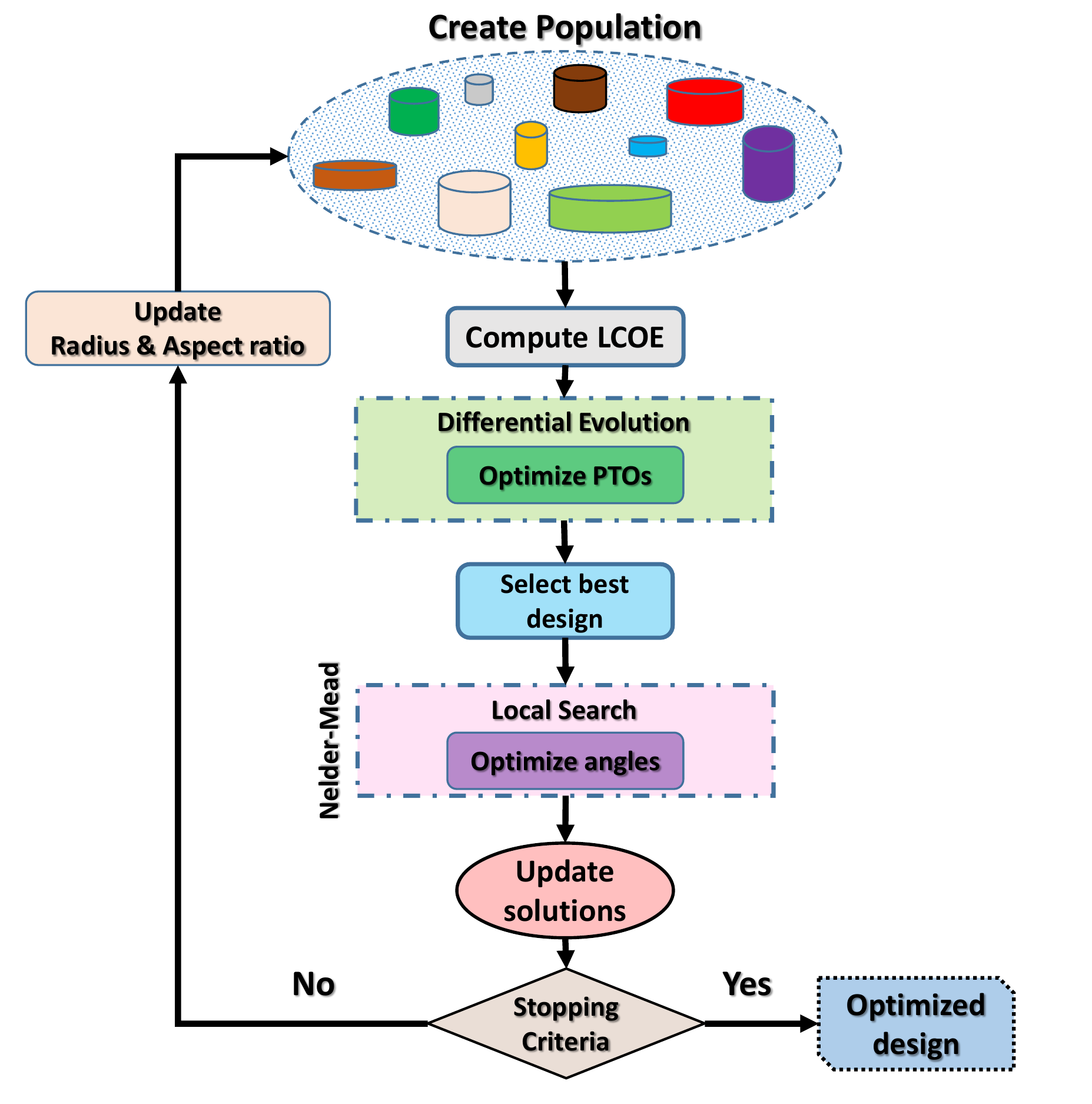}
    \caption{Scheme of proposed hybrid optimisation method (DE-NM).}
    \label{fig:op_method}
\end{figure}

As stated above, the optimised designs for this objective function converged to the assigned limits of the cylinder radius and aspect ratio. However, it would be interesting to see how sensitive the power output is to these two parameters. Therefore, for the next set of experiments, values of $a$ and $H/a$ are kept fixed and other parameters are optimised according to the procedure described in Fig.~\ref{fig:op_method}. A hybrid optimisation method DE-NM is implemented, where DE (Differential Evolution) is used to optimise the PTO parameters, while NM (Nelder-Mead) is used to optimise the tether angles. Due to the expensive computational cost of these experiments, DE-NM is run only once for each configuration (instead of ten times) in order to visualise the dependence of power on the buoy size.

\begin{figure}[hbt]
	\centering
	\includegraphics[width=0.7\linewidth]{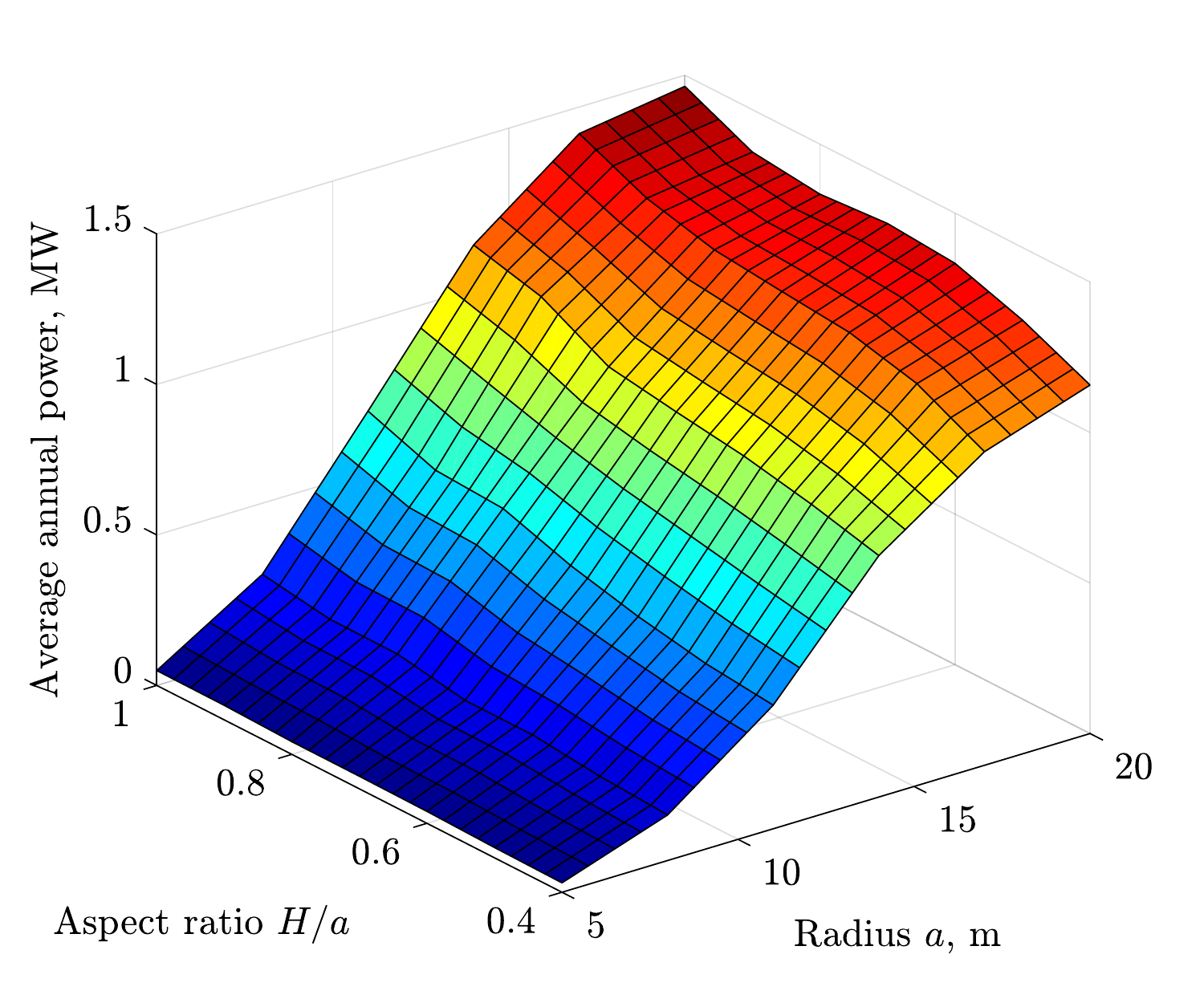}
	\caption{Average annual power output as a function of the buoy radius and aspect ratio obtained using DE-NM.}\label{fig:power_best_3D}
\end{figure}

As a result, Fig.~\ref{fig:power_best_3D} shows the annual average power production of the WEC as a function of its radius and aspect ratio. As expected, the power increases with increasing radius and height. However, a change in radius has a much stronger effect on the power output than a change in height. It is important to note that these results are obtained assuming linear wave theory. However, for such large buoys, the hydrodynamics are most likely nonlinear, and absolute values of power can vary, but the trend is expected to be the same.

\subsection*{Objective function $f_{O2}$}

\begin{figure}[hbt]
	\centering
	\includegraphics[width=0.5\linewidth]{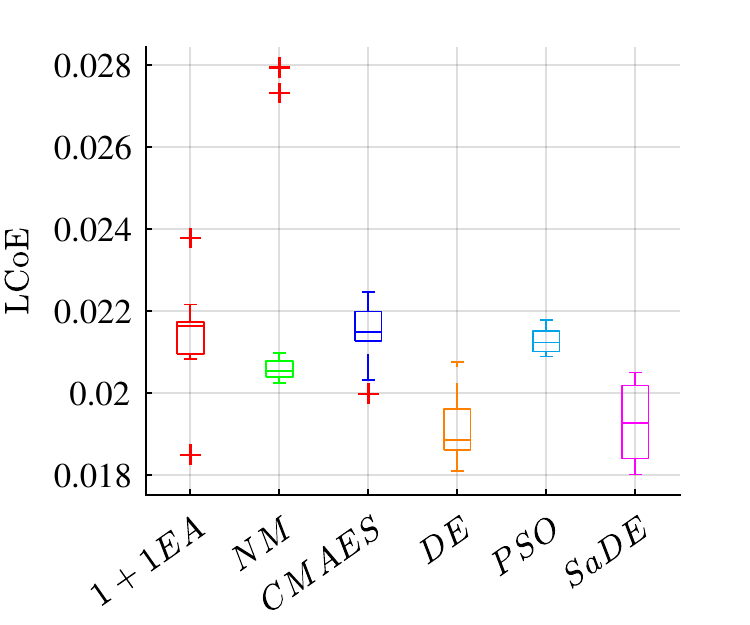}
	\caption{Performance comparison of various optimisation algorithms using objective function $f_{O2}$ (10 independent runs per each method).}\label{fig:max_lcoe}
\end{figure}

This objective function is related to the LCoE value approximated as a ratio of the generated energy to the significant mass of the system. The LCoE values for the best geometries obtained by our six optimisation methods are shown in Fig.~\ref{fig:max_lcoe}, and the corresponding design parameters are listed in Table~\ref{tab:min_lcoe}. Similar to the power maximisation study, all optimisation algorithms produced similar geometries with SaDE providing the lowest LCoE value. As expected, the WEC design with the lowest LCoE does not have the highest power output and it generates 680~kW.

\begin{table}[htb]
    \small
    \setlength\extrarowheight{-3pt}
    \caption{Design parameters when LCoE value is minimised.}
    \label{tab:min_lcoe}
    \centering
    \begin{tabular}{ccccccc}
        \toprule
       Parameter & 1+1EA & NM & CMA-ES & DE & PSO & SaDE \\
      \midrule
       $a$ [m]        & 14.7  & 14    & 12.2  & 12.5 & 14.1 & 13\\
       $H/a$        & 0.4   & 0.4   & 0.4   & 0.4   & 0.4 & 0.4\\
       $\alpha_t$ [deg]   & 16    & 17    & 22    & 18    & 21 & 18\\
       $\alpha_{ap}$ [deg] & 10   & 10    & 12    & 10    & 10 & 10\\
       \midrule
       LCoE$\times 10^3$    & 18.5 & 20.2 & 20 & 18.1 & 20.9 & 18\\
       Power [MW]    & 0.7 & 0.78 & 0.48 & 0.61 & 0.79 & 0.68\\
      \bottomrule
    \end{tabular}
\end{table}

The value of the buoy radius converges to 12-15 m which is very close to the announced design of CETO 6 system with a diameter of 25 m (radius is 12.5 m). This is a very interesting finding as the company run very expensive and accurate CFD simulations to analyse the potential power production, and used a reliable model to estimate the LCoE value. Therefore, it is surprising that the simplistic spectral-domain model with the approximate LCoE calculations generated the same buoy geometry.

In terms of the buoy height, all optimisation algorithms converged to the aspect ratio of 0.4 which is the lower assigned limit for this parameter. So the cylinder should be as short as possible (with a height of 4.8-6~m) in order to generate cheaper energy. A limit of 0.4 is set for the aspect ratio, since there is a possibility to install the power take-off machinery inside the buoy instead of having three independent PTO units on the seabed.

Due to the fact the $H/a \to 0.4$, the heave mode will be dominant in power production and the optimised tether angle $\alpha_t$ approaches values of 16-22 deg. Such values are also affected by the buoy weight and peak tether forces. So the smaller the tether angle, the less pretension force experienced by tethers and the smaller the mass of the anchoring system. Therefore, it is important to note that the buoy mass can also be included as an optimisation parameter in the future. It is not clear and requires more investigations why the tether attachment angle $\alpha_{ap}$ converged to its lower limit of 10 deg losing controllability over the pitch.

\begin{figure}[hbt]
	\centering
	\includegraphics[width=0.7\linewidth]{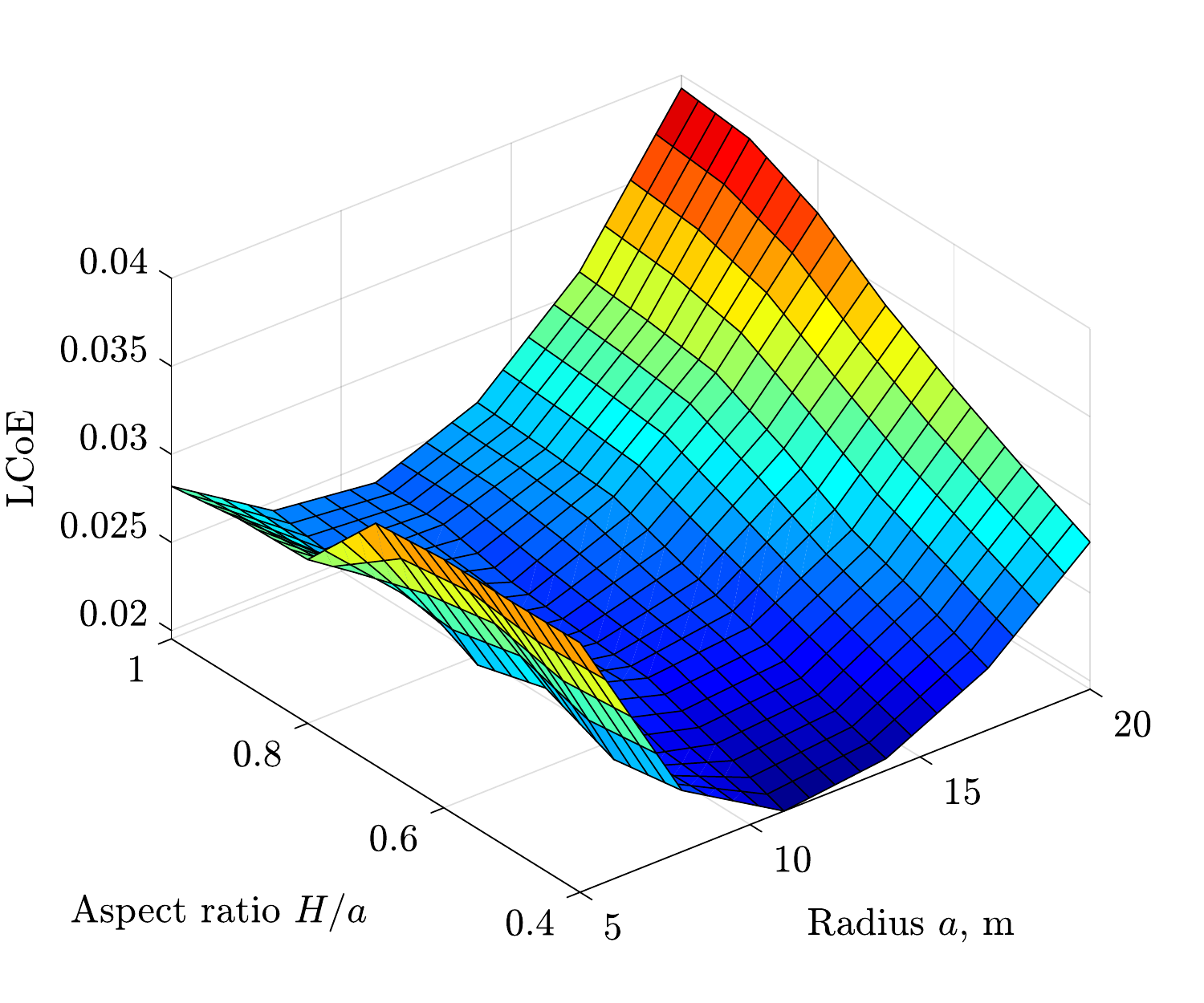}
	\caption{LCoE as a function of the buoy radius and aspect ratio obtained using DE-NM.}\label{fig:LCOE_best_3D}
\end{figure}

Similar to the power surface plot, the sensitivity of the approximated LCoE value to the buoy design parameters is shown in Fig.~\ref{fig:LCOE_best_3D}. It is obvious that energy cost increases with buoy height, while the minimum LCoE can be achieved with a buoy of 11-14~m radius regardless of its height.

\begin{figure}[t]
	\centering
	\includegraphics[width=0.5\linewidth]{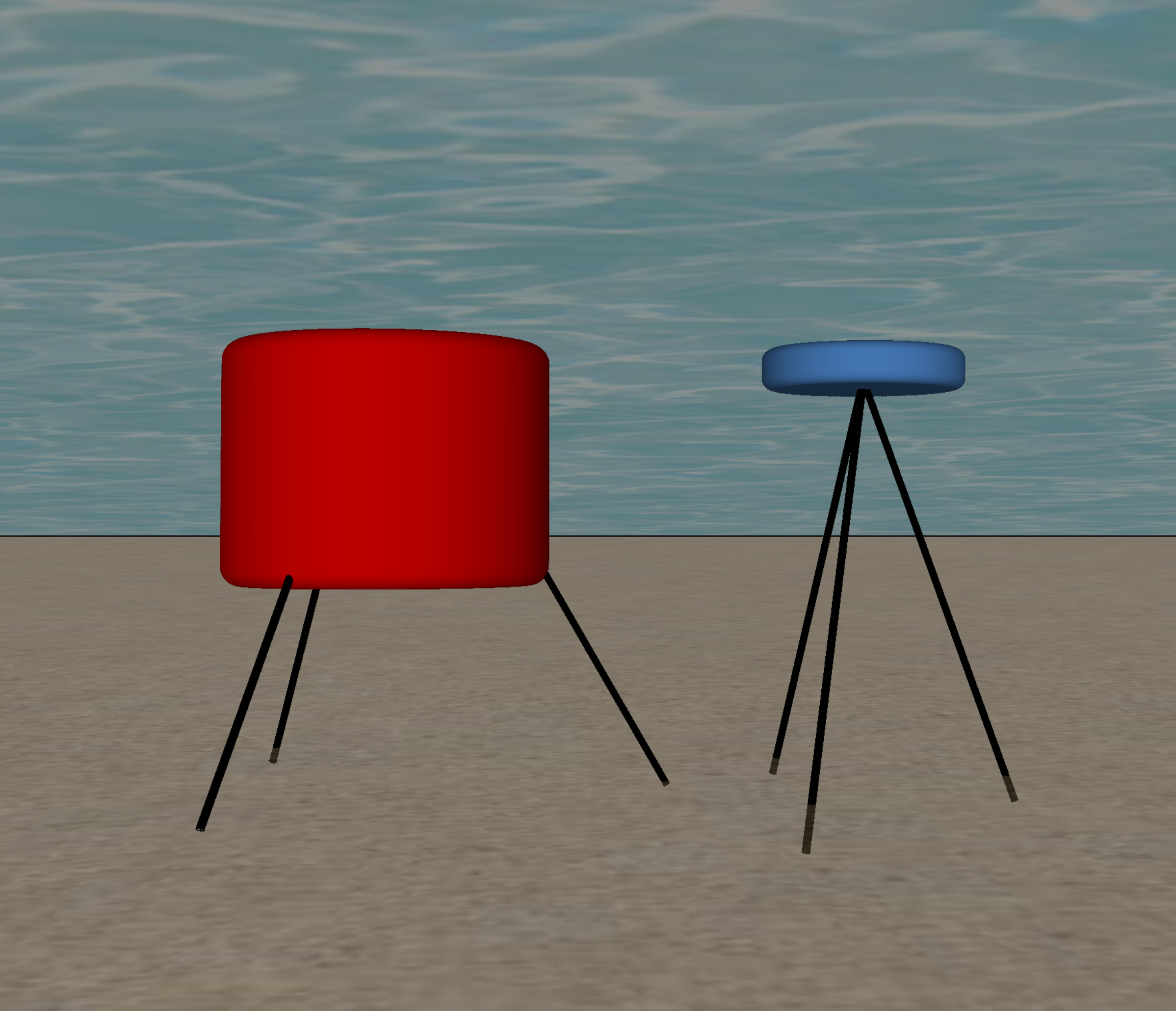}
	\caption{The best WEC geometries generated using two different objective functions. Left: power is maximised, right: LCoE value is minimised.}\label{fig:buoy_best}
\end{figure}

For comparison, the best two designs obtained for the two different objective functions are visualised in Fig.~\ref{fig:buoy_best}.

\subsection*{Discussion of optimisation methods}

\begin{figure}
\centering
\subfloat[]{
\includegraphics[clip,width=0.6\columnwidth]{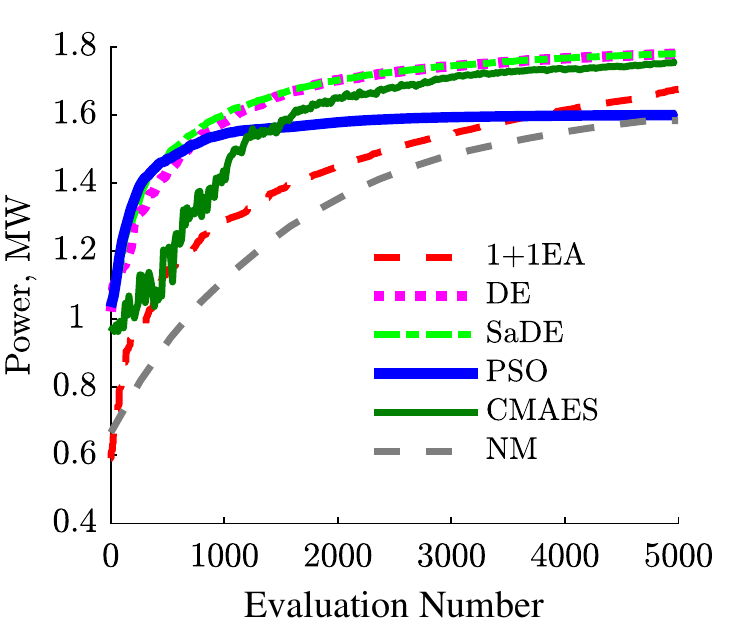}}\\
\subfloat[]{
\includegraphics[clip,width=0.6\columnwidth]{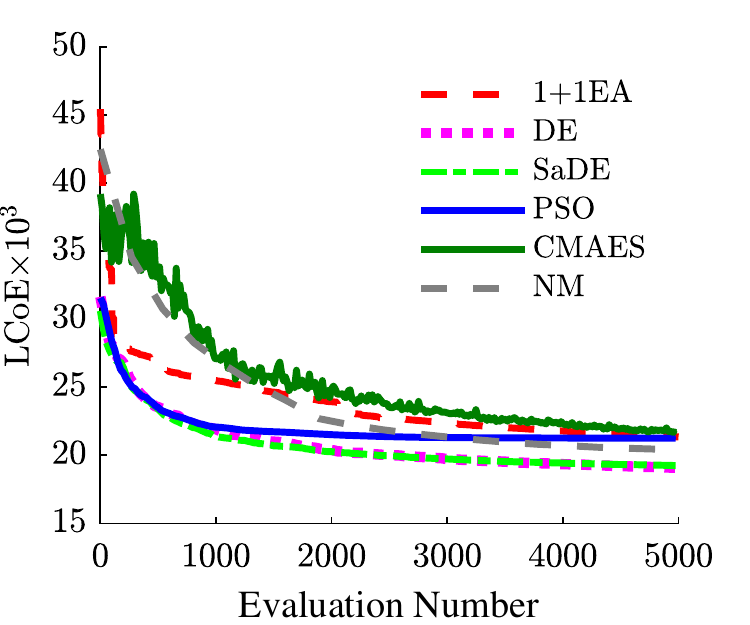}}
\caption{ Quality of configurations found over time using objective functions (a) $f_{O1}$, and (b) $f_{O2}$.}%
\label{fig:convergence}
\end{figure}

The results discussed so far are those found by the very end of the individual runs of the optimisation approaches. In the following, we briefly present how quickly the algorithms got there. To this end, Figure~\ref{fig:convergence} shows for both objective functions the convergence of the configurations to the final ones. 

Across both plots, the trends are largely comparable. DE and SaDE steadily make improvements and eventually perform best. PSO initially performs comparably, but converges quickly. This might be due to the existence of few local optima -- note that NM also performs badly in the case of power optimisation -- as our previous studies in \cite{neshat2018detailed, Neshat2019hybrid} indicated that certain aspects of the problem are unimodal or just bi-modal.\footnote{This means that there are either one or two local optima.} The self-adaptive CMA-ES performs well in the power optimisation scenario, however, it is among the worst performing then considering the LCoE -- this might again be explained by the bi-modality: CMA-ES typically works well for multi-modal problems, but the probability of making the jump here (from one corner of the search space to the other) is too small to be performed within only 5000 evaluations. Lastly, what the 1+1EA has in common with the others (except with PSO) is that all algorithms might be  able to benefit from a larger computational budget, as all of them have still been making small improvements even after evaluating almost 5000 configurations.

In summary, we conclude that the use of structurally different algorithms allowed us to mitigate the inherent search bias of the respective methods, also because the problem characteristics were not known beforehand. This way, we were able to find good configurations for both the power maximisation and the LCoE minimisation scenarios.


\section*{Acknowledgements}
 Leandro S. P. da Silva acknowledges the Australia-China Science and Research Fund, Australian Department of Industry, Innovation and Science; and the Adelaide Graduate Centre, the University of Adelaide. This work was supported with supercomputing resources provided by the Phoenix HPC service at the University of Adelaide.


\bibliographystyle{unsrt}  
\bibliography{sample-bibliography}
\end{document}